%% file: main.tex
\documentclass[10pt,twocolumn,letterpaper]{article}

\usepackage[pagenumbers]{cvpr} 

\usepackage{microtype}

\input{preamble}
\definecolor{cvprblue}{rgb}{0.21,0.49,0.74}
\usepackage[pagebackref,breaklinks,colorlinks,allcolors=cvprblue]{hyperref}
\usepackage{multirow}
\usepackage{booktabs}
\usepackage{hyperref}
\usepackage[table]{xcolor}

\def\paperID{*****} 
\def\confName{CVPR}
\def\confYear{2026}

\title{HiERO-StepG @ Ego4D Step Grounding Challenge:\\hierarchical activity understanding enables zero-shot step grounding}

\author{Andrea Zenotto\\
Politecnico di Torino\\
{\tt\small s327473@studenti.polito.it}
\and
Simone Alberto Peirone\\
Politecnico di Torino\\
{\tt\small simone.peirone@polito.it}
\and
Francesca Pistilli\\
Politecnico di Torino\\
{\tt\small francesca.pistilli@polito.it}
\and
Giuseppe Averta\\
Politecnico di Torino\\
{\tt\small giuseppe.averta@polito.it}
}

\begin{document}
\maketitle
\input{sec/0_abstract}    
\input{sec/1_intro}
\input{sec/2_method}
\input{sec/3_experiments}
\input{sec/4_conclusions}

{
    \small
    \bibliographystyle{ieeenat_fullname}
    \bibliography{main}
}


\end{document}

%% file: sec/0_abstract.tex
\begin{abstract}
Procedural activities follow well-defined structures: whether we consider a cooking recipe or a mechanic repairing a car, these activities naturally decompose in a hierarchy of steps and sub-steps.
Traditional approaches for step grounding require extensive annotations and scale poorly.
Instead, we argue that such hierarchical structure can emerge naturally from uncurated videos of human activities through recurring patterns of co-occurring actions and activities.
Our approach builds on HiERO, a weakly-supervised representation learning approach that maps close in the feature space actions that are functionally related to each other, leveraging only fine-grained action-level narrations.
In this feature space, procedure steps can be detected by a simple clustering, with no additional task-specific fine-tuning. 
For the Ego4D Step Grounding challenge, we augment this approach by ensuring fine and coarse level agreement in step assignments, enforcing strict temporal monotonicity of the grounded steps and post-processing the detected steps to reduce the impact of noisy predictions. 
We call this approach HiERO-StepG and it achieves \textbf{56.27\% on the R@1 (IoU = 0.3)} metric on the global leaderboard at submission time, ranking second while being completely zero-shot and not requiring procedure-specific annotations.
Project page: \href{https://github.com/andreazenotto/HiERO-StepG}{github.com/andreazenotto/HiERO-StepG}
\end{abstract}

%% file: sec/1_intro.tex
\section{Introduction}\label{sec:intro}

Several human activities have a strong procedural structure, as they  unfold though a hierarchy of steps and substeps. 
Cooking is a clear example, as recipes decompose into individual steps that must be executed in a precise order.
Procedure understanding covers a broad range of tasks, whose goal is to understand such hierarchical structure, reasoning about how steps and substeps interact and contribute toward the overall goal. 
\begin{figure*}[t]
    \centering
    \includegraphics[width=1\textwidth]{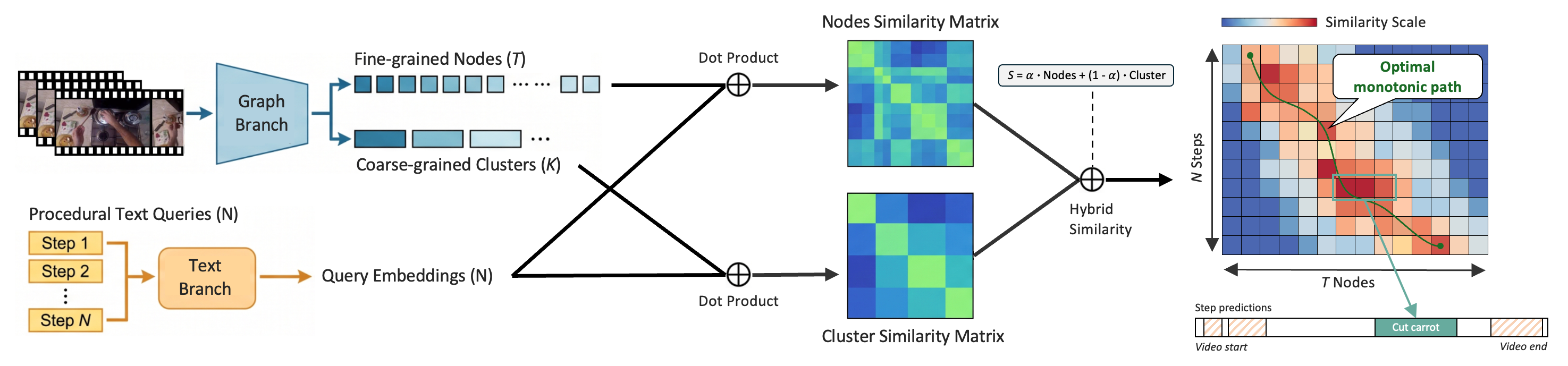}
    \caption{\textbf{Overview of the HiERO-StepG inference pipeline.} The video input is processed through a \textbf{Graph Branch} to extract fine-grained temporal nodes ($T$) and coarse-grained clusters ($K$), while $N$ procedural text queries are encoded via a \textbf{Text Branch}. Node- and cluster-level similarity matrices are computed via dot product with the textual queries and linearly combined using a weighting parameter $\alpha$ to construct the \textbf{Hybrid Similarity} matrix. Finally, a \textbf{Viterbi decoding} algorithm is applied to extract the optimal monotonic path, enforcing strict chronological ordering of the step assignments.}
    \label{fig:method}
\end{figure*}
Recently, Ego4D GoalStep~\cite{song2024ego4d} introduced a large scale set of annotations for Ego4D~\cite{ego4d}, with a hierarchical taxonomy of goal-oriented activity labels.
Here, we focus on the Step Grounding task~\cite{song2024ego4d} which aims at localizing the temporal boundaries of a given procedural step, given its free-form textual description. 
The task is closely related to \emph{moment localization}~\cite{zhang2020span}, where the goal is to temporally locate a textual query in a long video.
Previous works addressed this task from a fully supervised perspective, using annotated step boundaries to train a localization model~\cite{plou2024carlor,feng2025osgnet,pei2024egovideo}. 

Beyond relying on procedural annotations, procedural steps can emerge as action clusters across different videos depicting the same high-level activity~\cite{kukleva2019unsupervised,bansal2022my,chowdhuryopel}. 
However, these approaches assume that videos are from the same task and share the same steps, which limits generalization to unseen procedures.
StepFormer~\cite{dvornik2023stepformer} leverages a set of learnable queries to model the latent procedural steps in instructional videos, using only narrations as supervision. However, the learned representation captures procedures at a single temporal granularity, limiting its ability to model the hierarchical structure of human activities.
HiERO~\cite{peirone2025hiero} introduces a more flexible approach which learns an embedding space where distances between embeddings reflect the functional similarity of their corresponding narrations. 
Narrations provide a cheap form of supervision~\cite{lin2022learning}, enabling procedural steps to emerge from clusters of actions sharing similar narrations. HiERO is trained on approx. 3.8M clip-text pairs from EgoClip~\cite{lin2022egocentric} and exploits the natural co-occurrence of similar actions to learn context-aware embeddings.
In this feature space, steps and substeps emerge as clusters of actions sharing similar narrations, which enables zero-shot step~grounding. 

Here, we present HiERO-StepG which extends the zero-shot step grounding pipeline proposed in HiERO by imposing additional constraints in the grounding phase to: (i) ensure strict temporal monotonicity in the predicted steps, (ii) encourage assignments consistency across different temporal granularities, and (iii) adaptively refine the step boundaries to better capture the true temporal extent of the steps.
Unlike other approaches that leverage annotated step boundaries, our approach for step grounding is completely \emph{zero-shot} and has no access to any form of step annotations.
HiERO-StepG achieves 56.27\% on the R@1 (IoU = 0.3) metric on the global leaderboard at submission time, ranking second.

%% file: sec/2_method.tex
\section{Background: the HiERO architecture}\label{sec:method}
An input video $\mathcal{V}$ is first split into $T$ non-overlapping clips and clip-level embeddings are extracted from each clip using a video feature extractor (\textsc{LaViLa}~\cite{zhao2023learning}).
The video is then encoded as a video graph $\mathcal{G} = (\mathbf{X}, \mathcal{E}, \mathbf{p})$, where $\mathbf{X} \in \mathbb{R}^{T\times D}$ represents the node embeddings extracted from the video clips, edges $(i, j) \in \mathcal{E}$ connects the nodes $i$ and $j$ if their temporal distance is below a threshold $\tau$ and $\mathbf{p}$ encodes the temporal timestamp corresponding to each node. 
Each video is accompanied by an ordered list of narrations $\mathcal{N}=\{(n_i, t_i)\}$, where $n_i$ is a free-form fine-grained caption from the video at timestamp $t_i$.
The video graph $\mathcal{G}$ and its associated narrations $\mathcal{N}$ are then processed with a dual branch structure. 

The \emph{graph branch}~$\mathcal{F}_g$ maps the original graph to a set of L temporally coarsened graph representations: $\mathcal{G} \mapsto \{ \mathcal{G}^{(0)}, \mathcal{G}^{(1)}, \dots, \mathcal{G}^{(L)} \}$, obtained by halving the temporal resolution of the graphs at each stage.
This branch follows an encoder-decoder structure. The \emph{temporal encoder} implements local temporal reasoning, hierarchically aggregating information between close segments with progressively more coarse temporal granularity.
The \emph{function-aware decoder} extends temporal reasoning to nodes that may be temporally distant but functionally similar. 
Nodes are first clustered based on features similarity and the connectivity of the graph is updated to reflect the clusters. 
Spectral clustering~\cite{von2007tutorial} is used at this stage to model action clusters as densely connected regions of the graph based on their functional similarity (\emph{functional threads}).
Temporal reasoning is then performed inside each thread separately and the updated nodes are propagated back to the original graph.
The \emph{text branch} maps each narration to an embedding $\mathbf{n}_i \in \mathbb{R}^D$, that is aligned with node embeddings space through a video-text contrastive objective.

\noindent\textbf{Training and inference with HiERO.}
HiERO is trained with a video-text contrastive alignment objective $\mathcal{L}_{vna}$ that encourages the node embeddings to be close to the narration embeddings that fall within a predefined window around the node's timestamp. The functional threads clustering objective, $\mathcal{L}_{ft}$, promotes feature similarity among nodes assigned to the same cluster while pushing apart nodes belonging to different clusters.
HiERO is trained with a combination of the two objectives.
At inference time, procedural steps are obtained by clustering the node embeddings through spectral clustering, as done in the \emph{function-aware decoder}.

\section{The HiERO-StepG inference pipeline}\label{sec:method_inference}

We extend the zero-shot inference pipeline from HiERO to make the process more robust to textual ambiguities, overly generic step descriptions, and highly repetitive queries.
Specifically, we introduce a structured decoding process that ensures the correct temporal order of the grounded steps in the video. This approach is inspired by BayesianVSLNet~\cite{plou2024carlor}, which introduces a temporal prior to temporally shift the predictions of steps that appear more than once in the video.

\subsection{Feature extraction}
Given a video and a set of $n$ textual step queries, which are assumed to be temporally ordered according to their occurrence in the video, we proceed as follows.
First, we extract the input graph $\mathcal{G}$ from the video and compute the hierarchical graphs $\{ \mathcal{G}^{(0)}, \mathcal{G}^{(1)}, \dots, \mathcal{G}^{(L)} \}$. 
We then compute the cosine similarity between the node embeddings from the hierarchical graphs and the $\ell_2$-normalized query embeddings. 
To improve robustness against visual noise, we compute a \textbf{hybrid similarity} score by linearly combining the local similarity obtained from node embeddings in $\mathcal{G}^{(0)}$ with the global similarity from corresponding coarse cluster in $\mathcal{G}^{(L)}$ (Figure \ref{fig:method}). 
The resulting similarity matrix $S \in \mathbb{R}^{N \times T}$ stores in $s_{ij}$ the similarity score between the i-th query and the j-th temporal~node.

\subsection{In-order step grounding}
We then proceed assigning each segment from the video to a different step query, while enforcing a strict temporal ordering of the steps, \ie, step $i$ cannot occur before step $i-1$ has completed. 
We model this behavior by finding the optimal temporal sequence of nodes~$P = (p_1, p_2, \dots, p_N)$ that maximizes the joint similarity across all queries, subject to a strict monotonicity constraint:
$$ \max_{P} \sum_{i=1}^{N} S_{i, p_i} \quad \text{subject to} \quad t(p_{i-1}) \leq t(p_i) + \tau, $$
where $t(p_i)$ represents the timestamp (in seconds) at which step $i$ occurs, and $\tau$ is a minimal temporal tolerance window 
to account for slight overlap of the steps.
We then apply the \textbf{Viterbi decoding} algorithm to assign the steps.
Specifically, the algorithm iteratively builds the best assignment plan at time $t$ by considering the best plan up to $t-1$ and selecting the best assignment at time $t$. 
We define a score accumulation matrix~$D \in \mathbb{R}^{N \times T}$ and a backpointer matrix~$B \in \mathbb{N}^{N \times T}$.
The entry $D_{i, t}$ 
represents the maximum accumulated score obtainable by aligning the $i$-th query exactly to the $t$-th temporal node, provided that all preceding steps (from $1$ to $i-1$) have been optimally aligned, while $B$ tracks the optimal assignments.
For the first query ($i=1$), the best scores are given directly by the scores of the similarity map: $D_{1, t} = S_{1, t}$ for every $t$.
For each subsequent query $i$ and for each temporal node $t$, the score is calculated recursively by searching for the best node $t'$ to which the previous query ($i-1$) was assigned:
$$ D_{i, t} = S_{i, t} + \max_{t'} \left( D_{i-1, t'} + \phi(t', t) \right) $$
In this equation, the index $t'$ iterates over all possible previous temporal candidate nodes for query $i-1$. $\phi(t', t)$ is a temporal masking function that enforces the chronological constraint between the previous node $t'$ and the current node $t$. Specifically, $\phi(t', t) = 0$ if the transition is temporally valid ($t' \le t + \tau$), whereas $\phi(t', t) = -10^9$ otherwise.
This extreme penalty excludes non-monotonic paths from the solution space. 
At the end, we perform a backtracking pass to reconstruct the optimal path $P$. 

To expand the set of candidate steps, we propose a two-stage approach: point-level candidate generation followed by a dynamic, query-conditioned boundary expansion. 
In addition to the steps predicted through the Viterbi decoding algorithm, we extract additional candidates by taking the nodes with highest similarity in each candidate step and across the entire video. 
We then convert these candidate nodes into continuous segments by iteratively expanding their boundaries left and right until the local cosine similarity falls below a dynamic stopping threshold, defined as a relative fraction of the peak's maximum similarity. To capture varying temporal extents, we apply a predefined set of distinct expansion ratios to each peak, producing a concentric set of candidate segments. Finally, all segments are padded to guarantee a minimum duration, and an IoU-based NMS is applied to discard highly redundant segments, yielding the final Top-5 diverse predictions.

%% file: sec/3_experiments.tex
\section{Experiments}\label{sec:experiments}
\input{tables/challenge}
\input{tables/ablations}

\subsection{Implementation details}
We validate our approach on the Ego4D Step Grounding task~\cite{song2024ego4d}, using pre-extracted features from EgoVLP~\cite{lin2022egocentric} and \textsc{LaViLa}~\cite{zhao2023learning}. 
HiERO-StepG is trained on EgoClip~\cite{lin2022egocentric}, a curated subset of approx. 3.8M clip-text pairs sourced from Ego4D~\cite{ego4d}.
For the structured decoding and expansion phases, we set the hybrid similarity weighting parameter $\alpha = 0.7$ and the Viterbi temporal tolerance $\tau = 1.0$ second. To generate the concentric Top-5 predictions, we employ three relative expansion thresholds ($0.6$, $0.5$, and $0.4$), pad the segments to a minimum duration of $2$ seconds, and apply an NMS filter with an IoU threshold of $0.65$.
We report the results in terms of Recall at different Intersection over Union (IoU) thresholds, using R1@IoU=0.3 as the main metric.

\subsection{Challenge results}
We report the final results at submission time on the Step Grounding challenge in Table~\ref{tab:challenge}, showing significant improvements for all methods compared to the last year challenge.
Our approach, named \texttt{andreazenotto} on the global leaderboard ranks second, despite being completely \emph{zero-shot} not being specifically trained for the step grounding task.

\subsection{Ablations}

In the \textbf{HiERO baseline} configuration, each query step is grounded separately through spectral clustering of the fine-grained node embeddings. Each step is assigned to a candidate cluster based solely on cosine similarity, and the predicted temporal boundaries are strictly constrained to the start and end times of that specific cluster. To overcome the limitations of this approach, we evaluate our method by incrementally integrating the following proposed modules:

\begin{itemize}
    \item \textbf{Viterbi decoding}: tests the effectiveness of enforcing a globally coherent, monotonic temporal path for the step queries, compared to selecting the most similar segment for each query independently.
    \item \textbf{Hybrid similarity}: evaluates the impact of fusing fine-grained local similarities with coarse-level cluster predictions, rather than relying exclusively on the base level.
    \item \textbf{Query-conditioned expansion}: analyzes the benefit of adaptively refining segment boundaries outward from the peak nodes to handle local noise, as opposed to relying on static cluster boundaries.
    \item \textbf{Dynamic expansion}: measures the performance gain of generating a diverse, concentric set of predictions using multiple relative thresholds, rather than returning a single expanded window per grounded step.
\end{itemize}

We present a detailed comparison of all this variations in the following in Table~\ref{tab:ablation_results}, analyzing their impact on the step grounding~task.
The introduction of the temporal constraints through the \textbf{Viterbi decoding} algorithm brings a substantial improvement in performance ($+19.58\%$ on R1@IoU=0.3). Enforcing strict temporal ordering allows to better disambiguate between repeated steps across the videos and to ensure that the steps are predicted following the correct procedural order.
The \textbf{hybrid similarity} captures both local and global context from the ongoing activities, making the step assignments more robust to visual noise. This results in a $+5.72\%$ on R1@IoU=0.3 improvement when using the \textsc{LaViLa} features.

The \textbf{query-conditioned expansion} further improves localization performance by relaxing the predicted step boundaries. To prevent multiple candidate segments from collapsing into the same highly responsive temporal region, which would severely penalize Recall@5 metrics through redundant predictions, we also introduce an IoU-based Non-Maximum Suppression (\textbf{NMS}) step, ensuring that the top 5 predictions remain temporally diverse and do not excessively overlap.
These two modifications bring the R1@IoU=0.3 metric to $48.51\%$.
Overall, our inference pipeline significantly improves over the baseline on all the recall metrics.

%% file: tables/challenge.tex
\begin{table*}[htbp]
    \centering
    \scriptsize
    \caption{\textbf{Step Grounding performance on Ego4D Goal-Step~\cite{song2024ego4d} (test set)}. We report the results from the official leaderboard after the submission deadline. Our approach ranks second, despite being completely zero-shot. We report results for the \textsc{LaViLa} version of HiERO-StepG.}
    \label{tab:challenge}
    \begin{tabular}{llccccc}
        \toprule
        \textbf{Rank} & \textbf{Team} & Year & \textbf{R1 @ IoU = 0.30} & \textbf{R1 @ IoU = 0.50} & \textbf{R5 @ IoU = 0.30} & \textbf{R5 @ IoU = 0.50} \\
        \midrule
        1. & \texttt{yisen\_feng} & 2026 & 63.02 & 54.21 & 80.12 & 74.93 \\
        \rowcolor{gray!25} 2. & \texttt{andreazenotto (ours)} & 2026 & 56.27 & 40.20 & 77.39 & 61.38 \\
        3. & \texttt{kaname06} & 2026 & 53.39 & 45.43 & 79.91 & 73.59 \\
        4. & \texttt{yuki11} & 2026 & 52.41 & 44.55 & 79.09 & 72.64 \\
        5. & \texttt{willy06} & 2026 & 36.47 & 22.65 & 64.28 & 45.61 \\
        \midrule
        - & \texttt{OSGNet}~\cite{feng2025osgnet} & 2025 & 42.02 & 32.83 & - & - \\
        - & \texttt{BayesianVSLNet}~\cite{plou2024carlor} & 2025 & 35.18 & 20.49 & - & - \\
        - & \texttt{EgoVideo}~\cite{pei2024egovideo} & 2025 & 34.06 & 26.97 & - & - \\
        \bottomrule
    \end{tabular}
\end{table*}

%% file: tables/ablations.tex
\begin{table*}[htbp]
    \centering
    \scriptsize
    \caption{\textbf{Impact of different components on the Ego4D Goal-Step~\cite{song2024ego4d} validation split}. We compare two versions of our approach using 
    EgoVLP~\cite{lin2022egocentric} and \textsc{LaViLa}~\cite{zhao2023learning} embeddings.}
    \label{tab:ablation_results}

    \begin{subtable}[t]{0.49\textwidth}
        \centering
        \caption{EgoVLP features}
        \begin{tabular}{lcccc}
            \toprule
            \textbf{Model configuration} & \textbf{R1@0.3} & \textbf{R1@0.5} & \textbf{R5@0.3} & \textbf{R5@0.5} \\
            \midrule
            Raw features & 12.65 & 8.85 & 29.93 & 20.28 \\
            \midrule
            HiERO baseline & 13.98 & 9.89 & 33.38 & 23.08 \\
            \texttt{+} Viterbi decoding & 33.95 & 23.41 & 42.00 & 28.70 \\
            \texttt{+} Hybrid similarity & 39.63 & 27.40 & 45.95 & 31.39 \\
            \texttt{+} Query-cond. exp. + NMS & 45.40 & 32.33 & 62.51 & 46.99 \\
            \texttt{+} Dynamic expansion & \textbf{45.40} & \textbf{32.33} & \textbf{64.64} & \textbf{50.68} \\
            \bottomrule
        \end{tabular}
    \end{subtable}
    \hfill
    \begin{subtable}[t]{0.49\textwidth}
        \centering
        \caption{\textsc{LaViLa}-L features}
        \begin{tabular}{lcccc}
            \toprule
            \textbf{Model configuration} & \textbf{R1@0.3} & \textbf{R1@0.5} & \textbf{R5@0.3} & \textbf{R5@0.5} \\
            \midrule
            Raw features & 14.96 & 10.88 & 33.65 & 23.23 \\
            \midrule
            HiERO baseline & 15.70 & 11.27 & 36.37 & 25.44 \\
            \texttt{+} Viterbi decoding & 35.28 & 24.90 & 44.49 & 31.00 \\
            \texttt{+} Hybrid similarity & 41.00 & 28.81 & 48.65 & 33.74 \\
            \texttt{+} Query-cond. exp. + NMS & 48.51 & 34.58 & 65.06 & 49.34 \\
            \texttt{+} Dynamic expansion & \textbf{48.51} & \textbf{34.58} & \textbf{67.74} & \textbf{53.53} \\
            \bottomrule
        \end{tabular}
    \end{subtable}
\end{table*}

%% file: sec/4_conclusions.tex
\section{Conclusions}\label{sec:conclusions}
In this work, we presented HiERO-StepG, an inference pipeline specifically designed for zero-shot step grounding. Our approach demonstrates that procedural steps and substeps can emerge from the natural co-occurrence of similar actions in unscripted videos of human activities, without requiring procedure-specific annotations.
HiERO-StepG provides a robust pipeline for step grounding that ensures the grounded steps respect the correct procedural order, while accounting for visual noise in the videos. 
Our approach ranks second on the Ego4D Step Grounding challenge 2026 while being completely zero-shot. 